\documentclass[10pt,twocolumn,letterpaper]{article}

\usepackage{cvpr}
\usepackage{times}
\usepackage{epsfig}
\usepackage{graphicx}
\usepackage{amsmath}
\usepackage{amssymb}

\usepackage{algorithm}
\usepackage{algorithmic}
\usepackage[bold]{hhtensor}
\usepackage{tabularx}
\usepackage{multirow}

\usepackage[pagebackref=true,breaklinks=true,letterpaper=true,colorlinks,bookmarks=false]{hyperref}

\graphicspath{{./Figures/}}

\usepackage[breaklinks=true,bookmarks=false]{hyperref}

\cvprfinalcopy 

\begin{document}

\title{Learning to Search on Manifolds for 3D Pose Estimation of Articulated Objects}


\author{~~~~~~~~~~~~~~~Yu Zhang\\
\and
Chi Xu\\
\and
Li Cheng~~~~~~~~~~~~~~~~\\
\and
Bioinformatics Institute, A*STAR, Singapore \\
{\tt\small \{zhangyu, xuchi, chengli\}@bii.a-star.edu.sg}
}

\maketitle

\begin{abstract}
This paper focuses on the challenging problem of 3D pose estimation of a diverse spectrum of articulated objects from single depth images.
A novel structured prediction approach is considered, 
where 3D poses are represented as skeletal models that naturally operate on manifolds.
Given an input depth image, the problem of predicting the most proper articulation of underlying skeletal model is thus formulated as 
sequentially searching for the optimal skeletal configuration.
This is subsequently addressed by convolutional neural nets trained end-to-end to render sequential prediction of the joint locations as regressing a set of tangent vectors of the underlying manifolds.
Our approach is examined on various articulated objects including human hand, mouse, and fish benchmark datasets.
Empirically it is shown to deliver highly competitive performance with respect to the state-of-the-arts, while operating in real-time (over 30 FPS).
\end{abstract}

\section{Introduction}

3D Pose estimation of articulated objects is fundamental for many interesting 3D applications including skeleton-based object tracking~\cite{InteractingHands12,qian2014cvpr_realtimetracking},
action recognition~\cite{Mahasseni2016cvpr_lstmaction,Rahmani_2016_CVPR}, as well as higher-level behavior analysis~\cite{HuaKit:eccv14}.
Its goal is to accurately predict the joint locations of articulated objects such as human full-body~\cite{ShoEtAl:pami13}, human hand~\cite{tompson14tog}, and animals~\cite{xuchi2016ijcv}.
In this paper, we investigate the problem of 3D pose estimation of articulated objects from singe depth images.

It has been shown in e.g.~\cite{xuchi2016ijcv} that an articulated 3D object can be represented as a skeletal anatomy model in the form of a kinematic tree or chain. 
A pose or a specific articulation of the object can thus be characterized by sequence of rigid transformations (i.e. Lie groups),
resulting in a restricted set of possible joint locations that form the underlying pose manifolds~\cite{MurSasLi:book94}.
For example, a hand skeletal model may correspond to a kinematic tree that contains five sub-chains for five fingers all connected to the palm center joint as the root node.
Now, 3D location of each joint in our skeletal model can be predicted by a forward kinematic process with the rest joints fixed.
consequently an scan of the joints from root to finger tip nodes, together with corresponding rigid transformations collectively determine a pose.
In this paper, the pose estimation problem is formulated from the perspective of structured prediction~\cite{BakEtAl:book07},
in which the output space is the set of feasible joint locations defined by the underlying skeletal model.
This connection allows us to further exploit and extend the recent development in learning to search or L2S paradigm~\cite{DauEtAl:ml09,RosBag:arxiv14},
originally developed for structured prediction problems with discrete outputs, to address the problem of pose estimation in our context that involves continuous outputs.
Moreover, convolutional neural nets (CNNs) are employed as the L2S engine in our context,
which is referred to as deep-L2S that delivers very competitive performance in empirical evaluations.

\begin{figure*}
  \centering
  \includegraphics[width=0.97\textwidth]{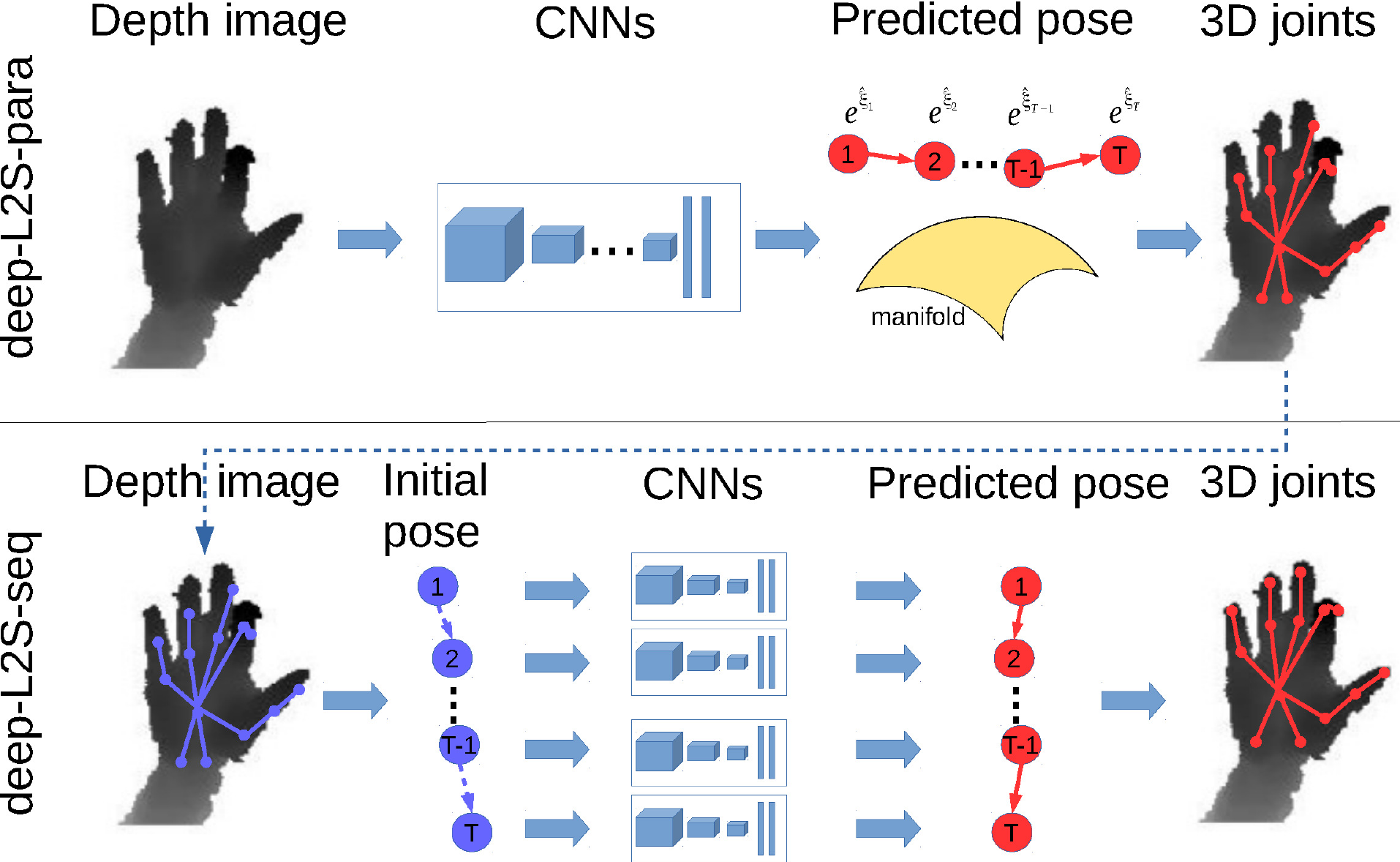}
   \caption{Pipeline of the two variants of our approach: Upper row illustrates our \emph{deep-L2S-para} variant that employs only one CNNs model to predict all joint locations at one time,
   while lower row presents our \emph{deep-L2S-seq} variant that works by executing multiple CNNs with each dedicating to a joint. See text for details.}
   \label{fig_liecnn}
\end{figure*}

Overall, our paper possesses three main contributions:
First, the pose estimation problem is explicitly formulated as L2S for structured prediction.
A dedicated CNN-based learning algorithm is further developed, which can be regarded as an extension of L2S that typically operates in discrete output spaces to continuous output spaces.
Second, a Lie-group output space is considered by working with 3D skeletal model.
As a result, our CNNs-based learning algorithm essentially directly maps from a raw depth image to points on manifolds expressed as rigid-transformation Lie-groups.
It is worth mentioning that to our knowledge this paper is among the first to learn CNNs with Lie-group valued outputs.
Third and last, our approach is empirically verified on diverse pose estimation applications, including human hand, mouse, and fish,
where it is shown to outperform the state-of-the-arts. 

A discussion of related works is presented below.
\vspace{-3mm}
\paragraph{3D pose estimation}

Enabled by the emerging commodity-level depth cameras,
recent efforts~\cite{ShoEtAl:pami13,XuChe:iccv13,tompson14tog,dahang2015iccv,Sun_2015_CVPR,Oberweger15a,Xingyi16ijcai,xuchi2016ijcv} have led to noticeable progress in 3D pose estimation.
Existing efforts can be roughly categorized into two groups, namely, regression-based and skeleton-based.
The regression-based methods regard the prediction of 3D joint locations directly as a multivariate regression problem in 3D Euclidean space.
The most known example is the 3D human full-body pose estimation used by Microsfot Kinect,
where, as elaborated in Shotton et al.~\cite{ShoEtAl:pami13}, random forests are engaged as the learning machine to regress joint locations.
Tompson et al.~\cite{tompson14tog} develop a dedicated CNNs for hand pose estimation, which is among the first application of deep learning for 3D pose estimation.

The skeleton-based methods, on the other hand, explicitly works with 3D skeletal models of the articulated objects.
This line of works is particularly dominant in hand pose estimation.
Once the hand region is located from the input depth image,
the work of~\cite{XuChe:iccv13} uses random forests to fit the induced 3D hand point cloud data by a proper articulation of its predefined hand skeletal model with fixed bone lengths, up to a global scaling factor.
Instead to estimate all joints at one time, Tang et al.~\cite{dahang2015iccv} propose a hierarchical sampling strategy for hand pose estimation, i.e.,
starting from its root joint and sequentially estimate the remaining joints of the skeleton.
Similarly, Xiao et al.~\cite{Sun_2015_CVPR} choose to first estimate the pose of hand palm and then estimate rest finger joints.
Oberweger et al.~\cite{Oberweger15a} utilizes CNNs to predict 3D joint locations, which is fed into a synthesizer to reconstruct hand depth images that are similar to the raw depth input.
Zhou et al.~\cite{Xingyi16ijcai} also incorporate physical joint constraints in that the feasible motion of each bone is confined by rotating around its joints.
Following the theory of Lie groups \& tangent vectors in robotics~\cite{MurSasLi:book94},
Xu et al.~\cite{xuchi2016ijcv} consider a random forests method that that operate on manifolds to predict good pose candidates based on skeleton models.

The most related work might be that of Xu et al.~\cite{xuchi2016ijcv}, where both consider working with skeletal models in terms of Lie groups as manifold-valued output spaces.
Different from~\cite{xuchi2016ijcv}, pose estimation is formulated in this paper as structured prediction where a L2S paradigm is specifically considered.
Moreover, CNNs are adopted to exploit the recent developments of the powerful deep learning machinery.
Note Lie groups have also been used for human body pose estimation~\cite{MikEtAl:ijcv03} based on inverse kinematics, which is replaced by CNNs predictors learned from training examples in our approach.
To our best knowledge, our work is the among the first such CNNs where the output space is defined as Lie groups.



\paragraph{Structured prediction and learning to search}


structured prediction~\cite{BakEtAl:book07} deals with structured outputs that goes beyond traditional binary classification problems.
Well-known examples include structured SVMs, CRFs, among others.
A critical fact here is that inference is also involved as a sub-module during the learning or training process.
Often the output space could be so large that renders exact inference intractable, thus we are left with only approximate inference.
However, most existing theory~\cite{BakEtAl:book07} works only with exact inference, which breaks with approximate inference.
It has since became an active research topic toward the interplay between approximate inference and learning, or the integration of learning and search.
Recently, the learning to search or L2S paradigm~\cite{DauEtAl:ml09,RosBag:arxiv14} has been developed to tackle this issue with promising results.
Here during training time and before learning starts, an optimal reference policy is obtained by fully exploiting labeled data.
In other words, training examples in this context can be viewed as expert demonstrations for sequential decision making.
%
It is also worth mentioning that there are also efforts from other perspectives.
For example, \cite{BelMcc:icml16} considers an a deep network architecture to learn an energy-based approximation of the true objective function that becomes tractable for inference.

\section{Our Approach}


In our approach, an articulated object is characterized by a 3D skeletal model in terms of its corresponding kinematic tree.
More specifically, our skeletal representation consists of a set of $T$ joints which are connected by a set of bones or link segments.
It is well known from robotics~\cite{MurSasLi:book94} that given a bone and by fixing one end of the bone as the base,
the position of the other end is determined by the group of rigid transformations in 3D Euclidean space,
which forms a Lie group that is usually referred to as the special Euclidean group or SE$(3)$, which is a differentiable manifold.
As such, an object pose can be defined as a particular articulation of the joints following its kinematic tree, that is,
a sequence of SE$(3)$ transformations or group actions from a home or identity pose.
As each of these SE$(3)$ group actions operates on a joint, together they corresponds to a product of Lie groups that is also a manifold.
These observations lead us to consider pose estimation as a structured prediction problem, which is further dealt with by our dedicated algorithm based on L2S paradigm and deep learning.
%
%
Prior to describe our algorithmic details, let us first provide a brief account of SE$(3)$ and its tangent vectors~\cite{MurSasLi:book94},
which are the basic building blocks in our skeletal model representation.
This is followed by a description of the L2S paradigm in our context. 

\paragraph{SE$(3)$, its tangent vectors, and the forward kinematics}

The Lie group SE$(3)$ is used in our skeleton models to depict the end joint position of a bone given its base joint is fixed.
It is defined as the set of rotational and translational transformations of the form $R \mathbf{x} + \mathbf{t}$ from the original position $\mathbf{x} \in \mathbb{R}^3$,
with $R$ any $3\times3$ rotational matrix and $\mathbf{t} \in \mathbb{R}^3$.
The tangent plane of SE$(3)$ at identity $\mathrm{I}$ forms its Lie algebra, $\mathrm{se}(3) := T_I \mathrm{SE}(3)$.
For any tangent vector $\hat{\vec{\xi}} \in \mathrm{se}(3)$, its algebraic form is
\begin{align}
\hat{\vec{\xi}} = \left(
    \begin{array}{cc}
        \hat{\omega} & \vec{\nu} \\
        \mathbf{0}^T & 0
    \end{array}
    \right),
\label{eq:xi_Hat}
\end{align}
where
\begin{align}
\hat{\vec{\omega}} = \left(
    \begin{array}{ccc}
        0 & -\omega_3 & \omega_2\\
        \omega_3 & 0 & -\omega_1\\
        -\omega_2 & \omega_1 & 0
    \end{array}
    \right),
\label{eq:omega_Hat}
\end{align}
and $\vec{\nu} \in \mathbb{R}^3$. We can further define $\vec{\omega}=(\omega_1, \omega_2, \omega_3)^T \in \mathbb{R}^3$.
Its corresponding Lie group action is realized by
\begin{align}
e^{\hat{\vec{\xi}}} = \begin{pmatrix} e^{\hat{\vec{\omega}}} & \vec{t} \\ \mathbf{0}^T & 1 \end{pmatrix} \in \mathrm{SE}(3),
\label{eq:LieGroupAction}
\end{align}
where $R \equiv e^{\hat{\vec{\omega}}}$ is the matrix exponential operation, and $\vec{t} = \vec{A}\vec{\nu}$ with the $3\times3$ matrix
\begin{align*}
\vec{A} = \vec{I} + \frac{\hat{\vec{\omega}}}{\|\vec{\omega}\|^2} \big( 1- \cos \|\vec{\omega}\| \big) + \frac{\hat{\vec{\omega}}^2}{\|\vec{\omega}\|^3} \big( \|\vec{\omega}\| - \sin \|\vec{\omega}\|  \big).
\end{align*}
It becomes clear that there is 6 degree-of-freedom (DoF) associated with each such group action, which can be equivalently represented as a 6-dim vector $\vec{\xi}=(\vec{\omega}^T, \vec{t}^T)^T$.


Without loss of generality, we consider a kinematic chain of $T^{'} \left( \leq T \right)$ joints.
Its forward kinematics from root up to the $t$-th joint can be naturally represented as the product of exponentials formula,
$g_{1:t} = e^{\hat{\vec{\xi}}_1 } e^{\hat{\vec{\xi}}_2 } \cdots e^{\hat{\vec{\xi}}_{t} }$, which holds true for $t=1,\cdots,T^{'}$.
Namely, for any $t$-th joint with home configuration being $\begin{pmatrix} \vec{x}_t \\ 1 \end{pmatrix}$, its new configuration is described by
\begin{align}
\label{eq_liecnn_forward}
\begin{pmatrix} \vec{x}_t^{'} \\ 1 \end{pmatrix} = g_{1:t} \begin{pmatrix} \vec{x}_t \\ 1 \end{pmatrix} = e^{\hat{\vec{\xi}}_1}e^{\hat{\vec{\xi}}_2}...e^{\hat{\vec{\xi}}_t} \begin{pmatrix} \vec{x}_t \\ 1 \end{pmatrix}, 
\end{align}
It becomes clear at this point that the set of these forward kinematics completely represents the entire object pose.
That is to say, the structured output space is sufficiently characterized by the set of tangent vectors of their specific joints, $\left\{ \vec{\xi}_t \right\}_{t=1}^T$.
It is also convenient to work with as locally their corresponding spaces can be regarded as Euclidean spaces.
Moreover, as illustrated in Fig.~\ref{fig_liecnn},
the transformation from such a tangent vector representation of the pose to its corresponding joint representation can be carried out
by applying Eq.~\eqref{eq_liecnn_forward} on individual joints following the kinematic tree.

\paragraph{Our deep-L2S paradigm}
We follow the notions of L2S paradigm~\cite{DauEtAl:ml09} and define the state space or structured output space $S$, action space $A$, and the policy space $\Pi$.
Here a state $s \in S$ corresponds to a particular pose, an action refers to a group action in Eq.\eqref{eq:LieGroupAction} applied on a joint, i.e. $e^{\hat{\vec{\xi}}} \in A$.
An agent follows a policy $\pi: S \rightarrow A$ to take an action $a \in A$, and arrives at a new state $s' \in S$ according a transition probability $p(s'|s,a)$.
At training time, due to the availability of ground-truth pose $\check{s} \in S$,
we are able to evaluate the incurred loss $L(\check{s}, s; \pi) \geq 0$ that measures the discrepancy between ground-truth $\check{s}$ and our prediction $s$. 
In particular, as loss can be decomposed to individual joints in our context, we consider
a tuple $\left(s_t, a_t, s_t', l_t(\check{s}_t, s_t'; \pi) \right)$ that describes the action-induced change of states restricted to a particular joint $t$.
Here $l_t$ denotes the instantaneous loss of joint $t$, which usually takes the form of vector norms, $\|g_{1:t}-y_t\|$,
where $g_{1:t}$ is the short-hand for $g_{1:t} \begin{pmatrix} \vec{x}_t \\ 1 \end{pmatrix}$, $y_t \in \mathbb{R}^3$ denotes the label, i.e. the ground-truth location of $t$-th joint.
Acumulatively we have $L(\check{s}, s; \pi) = \sum_t l_t (\check{s}_t, s_t; \pi)$.
In our context, a chain trajectory is a sequence of such tuples assembled from the root joint to the leaf joint of a kinematic chain.
A tree trajectory is then defined as a set of chain trajectories sharing the same root joint tuple, each focuses on a particular chain in our kinematic tree model.
Without loss of generality, assume the kinematics tree of interest containing $T$ nodes,
our goal is to learn a policy $\pi \in \Pi$ that minimizes the accumulative loss $L(\check{s}, s; \pi)$.
Specifically we consider in this paper a relatively restricted policy space where a policy always consists of a sequence of $T$ actions of fixed order, i.e. from root to leaf nodes.

By exploiting the deep learning methodology, CNNs are used to model the object pose as $\left\{ \vec{\xi}_t (\vec{w}_{t}) \right\}_{t=1}^T$.
To simplify the notation, we refer to $\vec{w}_{t}$ as the set of weights of a CNNs model with its $k$th element indexed as $w_{tk}$.
Following the fundamental back-propagation principle of deep learning, we have by the chain rule the partial derivative w.r.t. weight $w_{tk}$,
\begin{align}
\frac{\partial L}{\partial w_{tk}} = \frac{\partial}{\partial g_{1:t}} \|g_{1:t}-y_t\| \cdot \frac{\partial g_{1:t}(\vec{\xi}_t (w_{tk}))}{\partial w_{tk}}.
\label{eq:CNNs_BP}
\end{align}
Note here the first term of the RHS can be easily computed by the given norm.
The second term are a bit complicated as it involves manifold-based operations and the back-propagation process, which we will elaborate next.

Recall $\vec{\xi}$ is a 6-dim vector. For ease of notation, let us denote each of its $j$-th element as $\xi^{(j)}$ for $j=1, \cdots, 6$. In general, we have
\begin{align}
\label{eq_liecnn_bp}
&\frac{\partial g \left(\vec{\xi}_t (w_{tk}) \right)}{\partial w_{tk}} = \sum_j \frac{\partial g(\vec{\xi}_t)}{\partial \vec{\xi}_t^{(j)}} \cdot  \frac{\partial \vec{\xi}_t^{(j)}}{\partial w_{tk}} \\
&= \sum_j e^{\hat{\vec{\xi}}_1} e^{\hat{\vec{\xi}}_2}...e^{\hat{\vec{\xi}}_{t-1}} \frac{\partial e^{\hat{\vec{\xi}}_t}}{\partial \vec{\xi}_{t}^{(j)}} e^{\hat{\vec{\xi}}_{t+1}}...e^{\hat{\vec{\xi}}_T} \begin{pmatrix} \vec{x} \\ 1 \end{pmatrix} \cdot \frac{\partial  \vec{\xi}_{t}^{(j)}}{\partial w_{tk}}.  \nonumber\\
\nonumber
\end{align}
Here $\frac{\partial  \vec{\xi}_{t}^{(j)}}{\partial w_{tk}}$ follows the standard back-propagation rule in a neural nets with multivariate outputs.
For $\frac{\partial e^{\hat{\vec{\xi}}_t}}{\partial \vec{\xi}_{t}^{(j)}}$, consider e.g. $j=3$, we have $\vec{\xi}_{t}^{(3)} = \omega_3$.
By chain rule $\frac{\partial e^{\hat{\vec{\xi}}_t}}{\partial \vec{\xi}_{t}^{(3)}} = \frac{\partial \hat{\vec{\xi}}_t}{\partial \vec{\xi}_{t}^{(3)}} e^{\hat{\vec{\xi}}_t}$,
and from Eqs.~\eqref{eq:xi_Hat} and~\eqref{eq:omega_Hat} we have $\frac{\partial \hat{\vec{\xi}}_t}{\partial \vec{\xi}_{t}^{(3)}} = \begin{pmatrix} 0 & -1 & 0 & 0 \\ 1 & 0 & 0 & 0 \\ 0 & 0 & 0 & 0 \\ 0 & 0 & 0 & 0  \end{pmatrix}$,
this term $\frac{\partial e^{\hat{\vec{\xi}}_t}}{\partial \vec{\xi}_{t}^{(j)}}$ can thus be computed.

\paragraph{Two variants of our deep-L2S paradigm: \emph{deep-L2S-para} and \emph{deep-L2S-seq}}
Our deep-L2S paradigm opens doors to interesting explorations.
Here we consider two quite different variants: \emph{deep-L2S-para} and \emph{deep-L2S-seq}.
Our \emph{deep-L2S-para} variant considers to predict the entire structured output at one time, while \emph{deep-L2S-seq} proposes to predict each joint of the skeletal model one by one.
The corresponding algorithmic procedures are subsequently affected, as is illustrated in Fig.~\ref{fig_liecnn}:
In \emph{deep-L2S-para}, \emph{one} CNNs model is learned during the training stage. At test run, it is applied on the input object image to predict on the entire set of joints.

\begin{figure}
  \centering
  \includegraphics[width=0.45\textwidth]{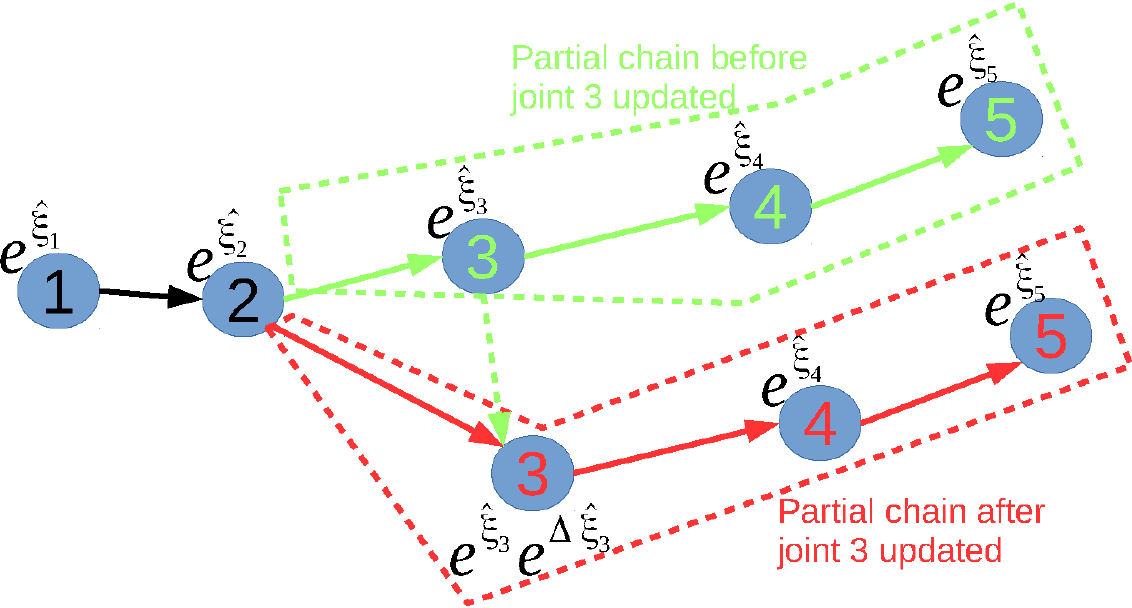}
  \caption{An illustration of the deep-L2S-seq method. This kinematic chain has 5 joints, with 1st joint being the root.
  Here it presents the moment when following the kinematic chain order to update the 3rd joint.}
  \label{fig_lie_single}
\end{figure}

In \emph{deep-L2S-seq}, on the other hand, $T$ distinct CNNs are trained, each is dedicated to one specific joint.
At test time, they are engaged \emph{one at a time} to incrementally predict on current joint based on the existing partial skeleton.
More specifically, let us consider a kinematic chain.
Here the specific articulation of the joints can be acquired by applying Eq.~\eqref{eq_liecnn_forward} on the home pose, which corresponds to the identity of the underlying differentiable manifold.
If additional group action say $e^{\Delta \hat{\vec{\xi}}_t}$ is performed on its $t$-th joint, it is known~\cite{MurSasLi:book94} that the forward kinematic process becomes
\begin{align}
\label{eq_liecnn_forward_1}
\begin{pmatrix} \vec{x}_t^{'} \\ 1 \end{pmatrix} = e^{\hat{\vec{\xi}}_1}e^{\hat{\vec{\xi}}_2}...e^{\hat{\vec{\xi}}_t} e^{\Delta \hat{\vec{\xi}}_t} \begin{pmatrix} \vec{x}_t \\ 1 \end{pmatrix},
\end{align}
Now learning in this context amounts to regress on the corresponding 6-dim vector $\Delta \vec{\xi}_t$ given everything else if fixed.
This is also illustrated in Fig.~\ref{fig_lie_single}.

\paragraph{Discussion on initial poses and the reference policies}
An initial pose is an important issue in our deep-L2S paradigm.
In this paper we consider the following set-up:
For \emph{deep-L2S-para}, the object's home pose is always considered as the initial pose.
Meanwhile, for \emph{deep-L2S-seq}, the predicted pose of \emph{deep-L2S-para} is employed as the initial pose,
which is established by applying Eq.~\eqref{eq_liecnn_forward} on the home pose.
Here the home pose plays the role of identity (e.g. zero w.r.t. to the real line).


In addition, following standard L2S, during training we need to secure a reference policy $\check{\pi}$.
Fortunately it can be directly obtained in our context: For each training example we have access to both the initial and the ground-truth poses of the input depth image,
the reference policy here boils to the sequence of group actions as in Eq.~\eqref{eq_liecnn_forward} that are executed on the initial pose to realize the ground-truth pose.

\section{Experiments}

\paragraph{Benchmark datasets}
Our approach is empirically examined on distinct objects including mouse~\cite{xuchi2016ijcv}, fish~\cite{xuchi2016ijcv},
and human hand~\cite{tompson14tog}, as is also illustrated in the respective panels of Fig.~\ref{fig_sample_imgs}.
%
%
%
More details are presented below:
\begin{itemize}
  \item  \emph{Mouse}~\cite{xuchi2016ijcv}: The training set has 104,096 depth images and test set has 4,125 depth images of different lab mice.
Depth images are captured with a top-mount Primesense Carmine depth camera, with image resolution being $640\times480$.
For each mouse image, its annotation contains 5 joints along the main spine that forms a kinematic chain.
  \item  \emph{Fish}~\cite{xuchi2016ijcv}: Training set contains 95,104 images, and test set contains 1,820 images of different zebrafish in water tanks,
  obtained via with a top-mount light field camera, and with image resolution $1,024\times1,024$.
21 3D joints along the main spine is annotated for each fish.

  \item  \emph{Human hand}~\cite{tompson14tog}: It consists of 72,757 training and 8,252 test images.
36 3D joints are annotated, while following~\cite{oberweger15}, 14 of them are in use.
The hand depth images considered are acquired from a front-view MS Kinect camera, with a resolution of $640\times480$.
\end{itemize}

\begin{figure}
  \centering
  \includegraphics[width=0.475\textwidth]{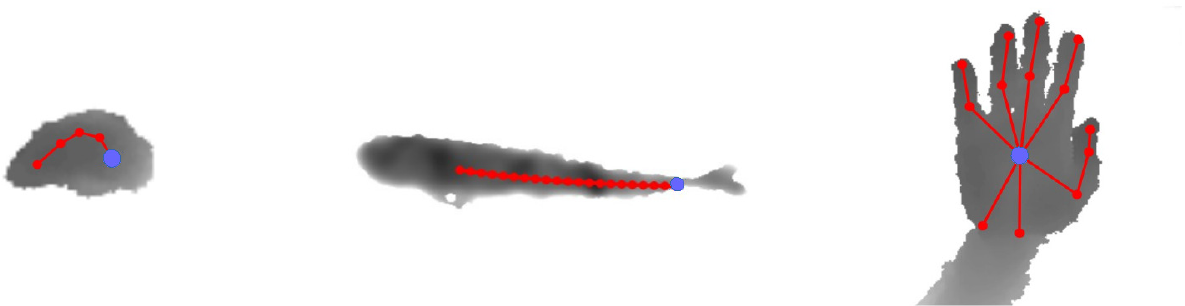}
  \caption{A portry of the skeletal models for mouse, fish, and human hand applications, respectively.
  The colored points represent the corresponding joints, while the blue point specifically denotes the root joint of the kinematic tree (chain).}
  \label{fig_sample_imgs}
\end{figure}

\begin{table*}
 \caption{A summary of the input patch sizes for \emph{CNN-baseline} and the two variants of our approach when operating on different benchmarks.}
 \label{table_cnn_strucutres}
 \centering
 \begin{tabular}{|l|l|l|l|}
  \hline
  Benchmarks vs. Methods   &  CNN-baseline  &  deep-L2S-para &   deep-L2S-seq \\
  \hline
  Mouse      & $80\times 80$, Table~\ref{table_mouse_cnn_baselie} & $80\times 80$, Table~\ref{table_mouse_cnn_baselie}	& $40\times 40$, Table~\ref{table_mouse_cnn_sequential}	\\
  Fish      & $224\times 224$, AlexNet~\cite{NIPS2012alexnet} 	& $224\times 224$, AlexNet~\cite{NIPS2012alexnet} 	& $100\times 100$, Table~\ref{table_nyu_hand_cnn} 	\\
  Human hand      & $128\times 128$, Table~\ref{table_nyu_hand_cnn}	& $128\times 128$, Table~\ref{table_nyu_hand_cnn}	& $100\times 100$, Table~\ref{table_nyu_hand_cnn}	\\
  \hline
 \end{tabular}
\end{table*}

\paragraph{Preprocessing, network architectures, and comparison methods}
Our approach empirically requires some basic preprocessing steps as follows.
Since the object of interest often covers only a small portion of a depth image, it is natural to consider detecting and extracting the square-shaped patch where the object resides.
For for mouse and fish applications, we directly follows the simple image processing process as specified in~\cite{xuchi2016ijcv}.
For human hand, the hand patches from the NYU benchmark are obtained using the code released by~\cite{oberweger15}.
Now each input depth patch contains the entire object, which can also be considered as a 3D point cloud.
We follows~\cite{oberweger15} to normalize each input point cloud individually by rescaling to prescribed size as summarized in Table~\ref{table_cnn_strucutres}
with their empirical centroid as origin.

Besides the two variants of our approach,
we consider the direct application of AlexNet CNNs~\cite{NIPS2012alexnet} in our context, which we refer to as \emph{CNN-baseline}.
This works well for fish dataset, it leads to poor performance for the rest two benchmarks, due to the significant variability across these datasets.
Empirically better performance is obtained when working with reduced input size. Similar observations are also found for our two variants.
Table~\ref{table_cnn_strucutres} summarizes the input size we have considered for dedicated situations.
Since the architecture of a CNNs hinges on the input size,
we also list specific network architectures in Tables~\ref{table_mouse_cnn_baselie},~\ref{table_mouse_cnn_sequential}, and~\ref{table_nyu_hand_cnn},
and linked them up in Table~\ref{table_cnn_strucutres} as the entry point.
Here AlexNet~\cite{NIPS2012alexnet} in our tables means the same network architecture of AlexNet~\cite{NIPS2012alexnet} is employed.
The notations of Conv., Pool, and Fc. in our tables refer to filter sizes considered in convolutional layer and max-pooling layer, as well as length of fully connected layer, respectively.
More specifically, we do not consider the usage of max-pooling layers when working with mouse benchmark.
This is mostly due to the relatively small input patches of mice where we would rather concern on the loss of fine details.
Note in particular that differing from our \emph{deep-L2S-para} variant that operates on the input patch of the entire object of interest,
in \emph{deep-L2S-seq}, to regress on current joint, we instead work with a local patch that centers around its immediate previous joint.
This partly explains the relative small patch size considered by \emph{deep-L2S-seq} as presented in Table~\ref{table_cnn_strucutres}.

We also compare with the recent state-of-the-art methods of these benchmarks,
including \emph{Lie-X} of~\cite{xuchi2016ijcv}, \emph{Deep prior} of~\cite{oberweger15} on mouse and fish benchmarks,
as well as \emph{CNNs} of~\cite{tompson14tog}, \emph{Deep model} of~\cite{Xingyi16ijcai}, and \emph{Feedback loop} of~\cite{Oberweger15a}.


Throughout experiments, the usual $l2$-norm is used for the vector norm considered in Eq.~\eqref{eq:CNNs_BP}.
For performance evaluation, the commonly used metric of \emph{average joint error} is adopted, which computes the averaged Euclidean distance over all 3D joints.
Our experiments are carried out on a standard desktop computer with an Intel Core i7 CPU, 32 GB of RAM, as well as a nVidia Titan-X GPU.
Our implementation is Matlab-based, and uses the MatConvNet library.
In terms of runtime speed (FPS), when working with mouse, fish, and hand benchmarks, the speed of \emph{deep-L2S-para} is 190.2, 103.1, and 161.2, respectively,
while the speed of \emph{deep-L2S-seq} is 176.2, 31.7, and 35.5, respectively. In summary, our approach is capable of realtime applications.

\begin{table}
\scriptsize
 \centering
 \caption{The network architecture considered on mouse benchmark that is adopted by \emph{CNN-baseline} and \emph{deep-L2S-para}.}
 \label{table_mouse_cnn_baselie}
 \begin{tabular}{|c|}
  \hline
 Input	\\
 \hline
 Conv. $7\times7\times32$, stride 2 \\
 \hline
 ReLU  \\
 \hline
 Conv. $7\times7\times128$, stride 1 \\
 \hline
 ReLU \\
 \hline
 Fc. 2048, ReLU   \\
 \hline
 Fc. 128	\\ 	
 \hline
 Output	\\
  \hline
 \end{tabular}

 \caption{The network architecture considered on mouse benchmark that is adopted by \emph{deep-L2S-seq}.}
 \label{table_mouse_cnn_sequential}
 \begin{tabular}{|c|}
  \hline
  Input \\
  \hline
  Conv. $3\times~3\times~16$, stride 1\\
  \hline
  ReLU  \\
  \hline
  Fc. 1024, ReLU	\\
  \hline
  Fc. 128 \\
  \hline
  Output \\
  \hline
 \end{tabular}

 \caption{The network architecture considered on human hand benchmark that is adopted by \emph{CNN-baseline} \emph{deep-L2S-para}, and \emph{deep-L2S-seq}.
 It is also employed by \emph{deep-L2S-seq} on fish benchmark. 
 }
 \label{table_nyu_hand_cnn}
 \begin{tabular}{|c|}
  \hline
  Input   \\
  \hline
  Conv. $7\times7\times128$, stride 1 \\
  \hline
  Pool $4\times4$, stride 2\\
  \hline
  ReLU \\
  \hline
  Conv. $7\times7\times128$, stride 1 \\
  \hline
  Pool $4\times4$, stride 2\\
  \hline
  ReLU \\
  \hline
  Conv. $7\times7\times128$, stride 1 \\
  \hline
  Pool $4\times4$, stride 2\\
  \hline
  ReLU \\
  \hline
  Fc. 2048, ReLU \\
  \hline
  Fc. 128  \\
  \hline
  Output \\
  \hline
 \end{tabular}

\end{table}

\paragraph{Results on mouse and fish benchmarks}

\begin{figure}
  \centering
  \includegraphics[width=0.47\textwidth]{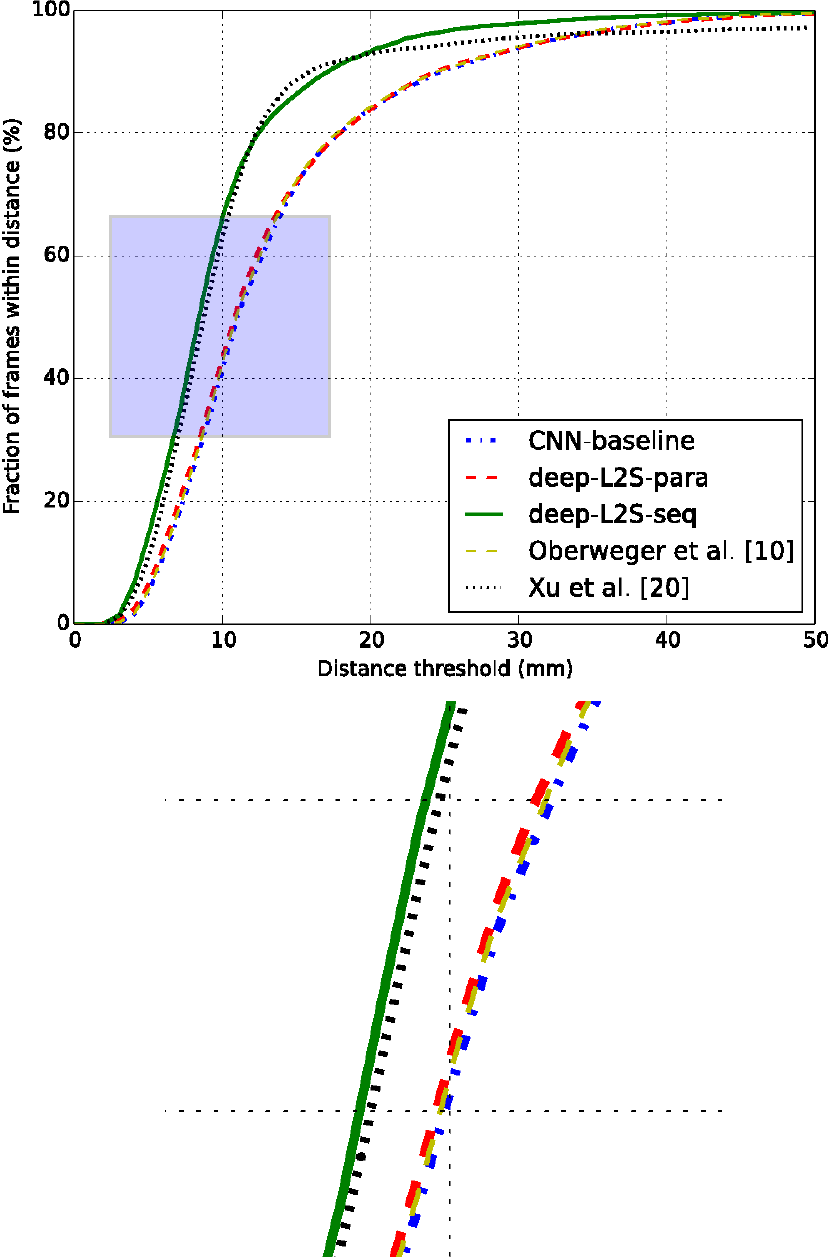}
  \caption{Cumulative error distributions of comparison methods on mouse benchmark of~\cite{xuchi2016ijcv}. The bottom figure is zoomed in from the blue region in the upper figure. Best viewed in color.}
  \label{fig_mouse_cum}
\end{figure}

\begin{figure}
  \centering
  \includegraphics[width=0.47\textwidth]{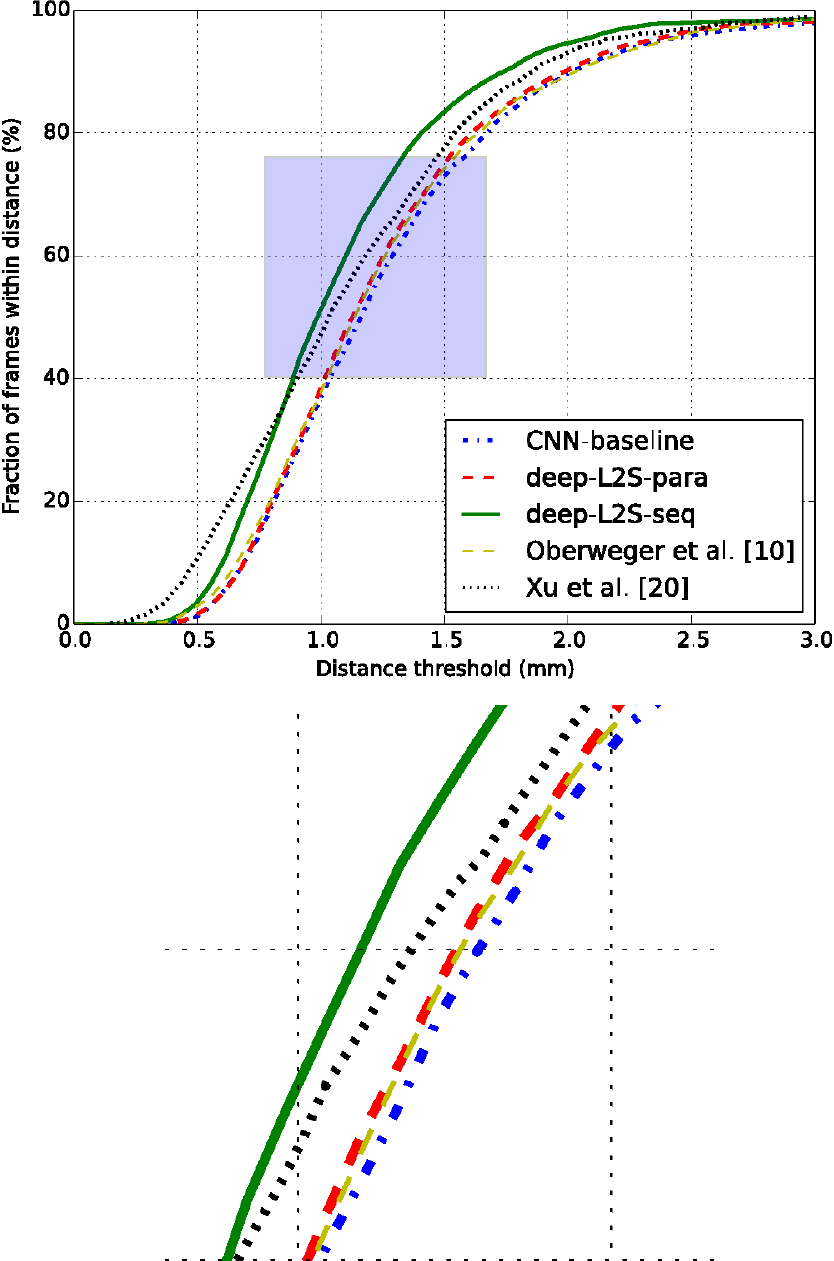}
  \caption{Cumulative error distributions of comparison methods on fish benchmark of~\cite{xuchi2016ijcv}.  The bottom figure is zoomed in from the blue region in the upper figure. Best viewed in color.}
  \label{fig_fish_cum}
\end{figure}

Comparison methods are first evaluated on mouse and fish applications that contain one kinematic chain as illustrated in Fig.~\ref{fig_sample_imgs}.
The average joint error is reported for mouse and fish in Table~\ref{table_mouse_err} and Table~\ref{table_fish_err}, respectively.
Furthermore, their cumulative error distributions are presented in Fig.~\ref{fig_mouse_cum} and Fig.~\ref{fig_fish_cum}, respectively.

In general, the \emph{CNN-baseline} method, \emph{deep prior}~\cite{oberweger15}, and our \emph{deep-L2S-para} variant produce similar performance for both applications.
They are further overtaken by \emph{Lie-X}~\cite{xuchi2016ijcv} and our \emph{deep-L2S-seq} variant by a noticeable margin.
Overall our \emph{deep-L2S-seq} outperforms \emph{Lie-X}, and works especially well during the presence of large errors (0.8-2.5 mm for fish and 20-50 mm for mouse).
These trends are clearly visible from the cumulative error curves of Fig.~\ref{fig_mouse_cum} and Fig.~\ref{fig_fish_cum}.
Note for mouse benchmark in particular, as in Fig.~\ref{fig_mouse_cum}, \emph{Lie-X} seems to incur noticeable amount of extremely large errors of over 50 mm.
In comparison, our \emph{deep-L2S-seq} gracefully goes to almost 100\% when error is around 40-45 mm.
Another observation is by feeding with the output pose of \emph{deep-L2S-para} as initial pose guess,
empirically our \emph{deep-L2S-seq} always outperforms the initial guess a lot: Over 1.5 mm for mouse and around 0.1 mm for fish.

\begin{figure*}
  \centering
  \includegraphics[width=0.98\textwidth]{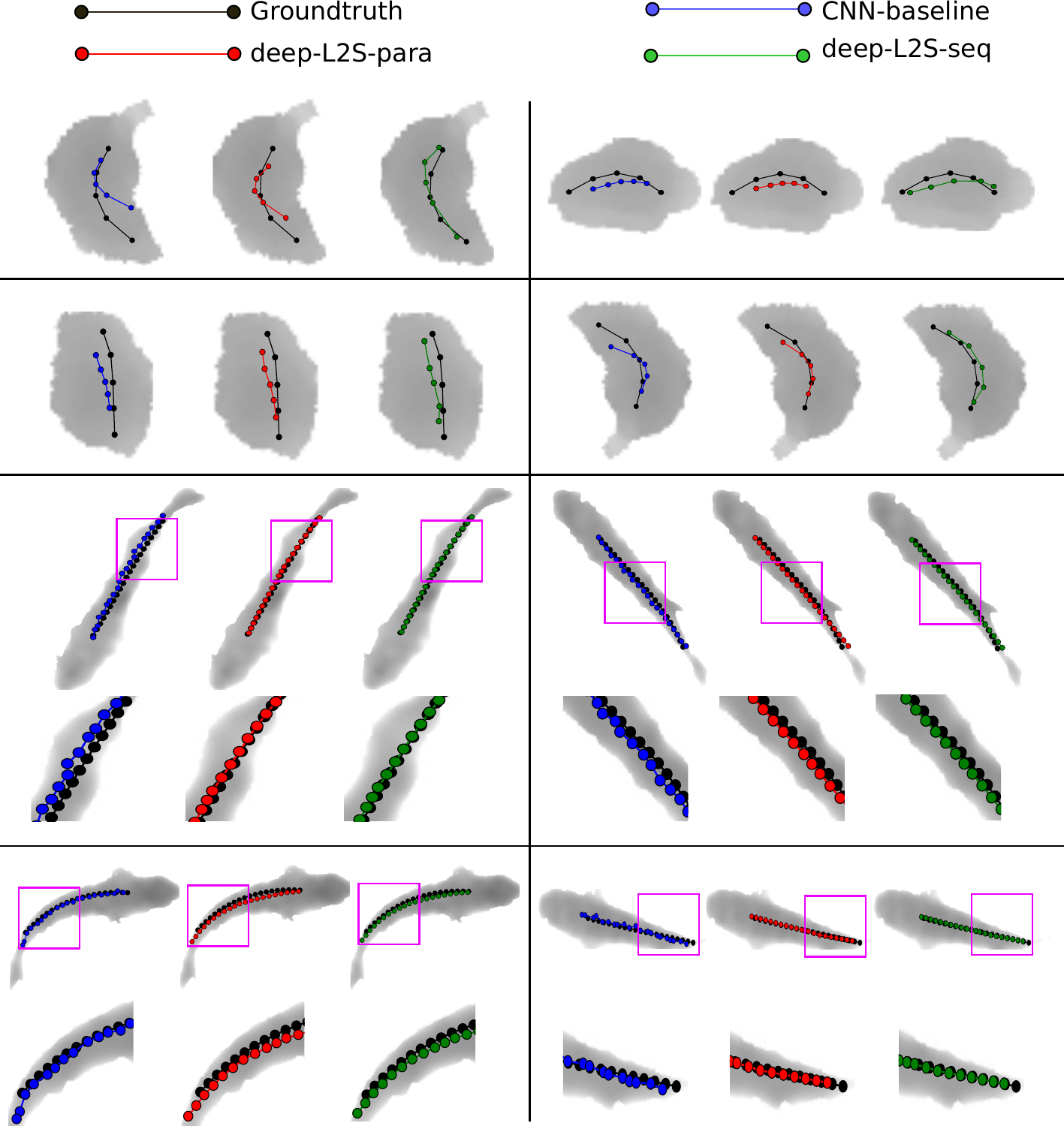}
  \caption{Visual comparisons of performance on mouse and fish benchmarks.
  Each group presents results on \emph{CNN-baseline} (blue), \emph{deep-L2S-para} (red), and \emph{deep-L2S-seq} (green), respectively.
  Black curve denotes the ground-truth. Zoom-in local regions are shown at the bottom of each image to highlight the difference for fish. Best viewed in color.}
  \label{fig_visual_mousefish}
\end{figure*}

Fig.~\ref{fig_visual_mousefish} displays visual comparisons. 
On the mouse dataset, we can see a clear difference among the three methods in comparison, where the predicted skeletons from \emph{deep-L2S-seq} are clearly closer to that of the ground-truths.
On the fish dataset, the difference among these methods is relatively less prominent. We can still see the improvement of our skeleton based methods over \emph{CNN-baseline}.
It is worth noting that sometimes the predicted tail joints of \emph{CNN-baseline} are not well aligned with the skeleton,
while our \emph{deep-L2S-seq} delivers consistently satisfactory results that are visually more appealing.

\begin{table}
 \caption{Average joint error on mouse benchmark.}
 \label{table_mouse_err}
 \centering
 \begin{tabular}{|l|l|c|}
  \hline
  \multicolumn{2}{|l|}{Methods }  & Error (mm)  \\
  \hline
  \multicolumn{2}{|l|}{Lie-X~\cite{xuchi2016ijcv}} &	6.67	\\
  \multicolumn{2}{|l|}{Deep prior~\cite{oberweger15}} &	8.14 \\
  \multicolumn{2}{|l|}{CNN-baseline} &	8.24 \\
  \hline
  \multirow{2}{*}{ours}
  & deep-L2S-para  &  7.92 \\
  & deep-L2S-seq  &  \textbf{6.35}   \\
  \hline
 \end{tabular}
\end{table}

\begin{table}
 \caption{Average joint error on fish benchmark.}
 \label{table_fish_err}
 \centering
 \begin{tabular}{|l|l|c|}
  \hline
  \multicolumn{2}{|l|}{Methods }    & Error (mm)   \\
  \hline
  \multicolumn{2}{|l|}{Lie-X~\cite{xuchi2016ijcv}} &	0.68	\\
  \multicolumn{2}{|l|}{Deep prior~\cite{oberweger15}} & 0.79	\\
  \multicolumn{2}{|l|}{CNN-baseline} & 0.78	\\
  \hline
  \multirow{2}{*}{ours}
  & deep-L2S-para  &  0.75   \\
  & deep-L2S-seq  &  \textbf{0.66}   \\
  \hline
 \end{tabular}
\end{table}

\begin{figure*}
  \centering
  \includegraphics[width=0.98\textwidth]{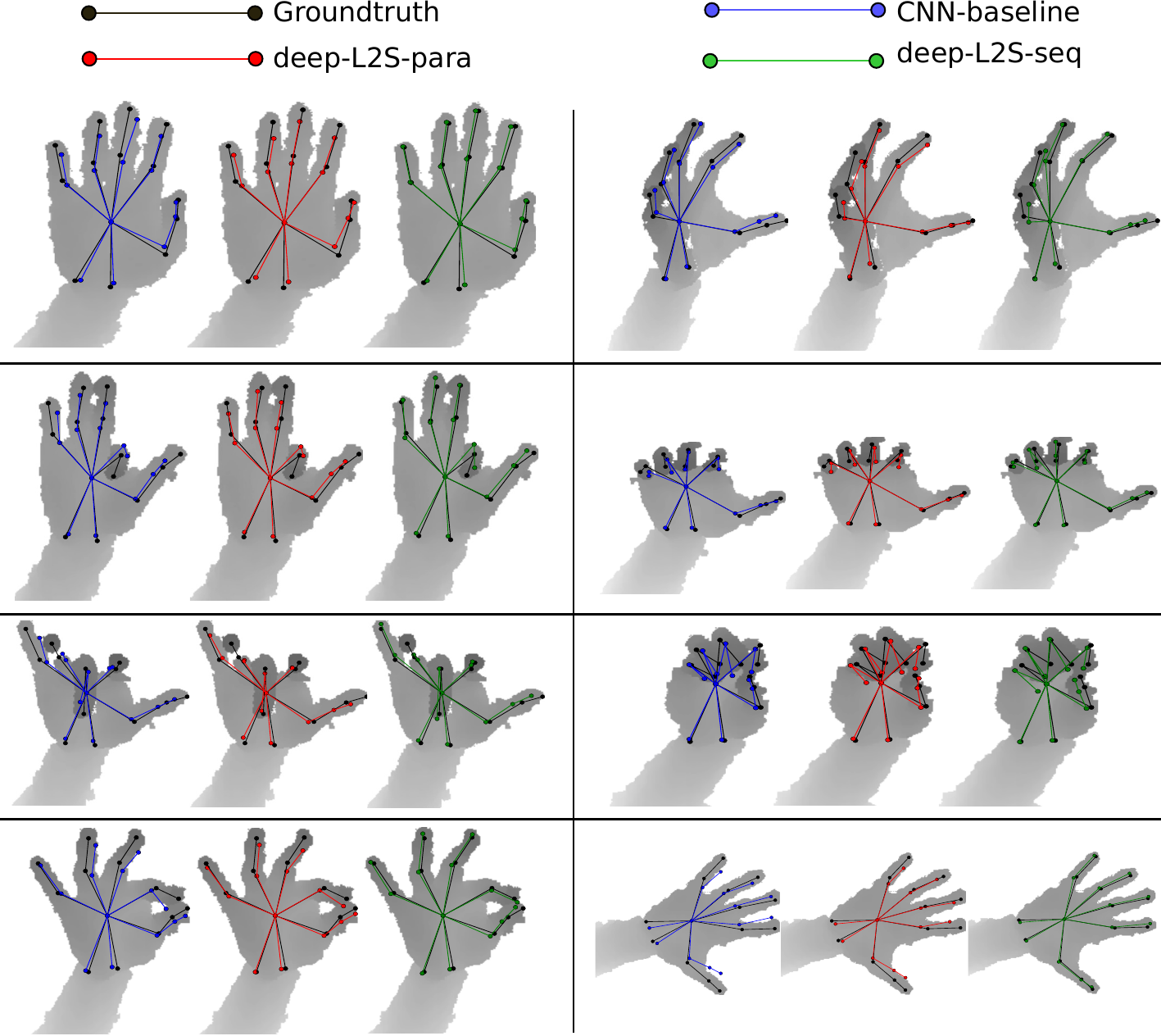}
  \caption{Visual comparisons of performance on hand benchmarks.
  Each group presents results on \emph{CNN-baseline} (blue), \emph{deep-L2S-para} (red), and \emph{deep-L2S-seq} (green), respectively.
  Black curve denotes the ground-truth. Best viewed in color.}
  \label{fig_visual_hand}
\end{figure*}

\begin{table}
 \caption{Average joint error on human hand benchmark~\cite{tompson14tog} based on 14 joints following~\cite{oberweger15}).}
 \label{table_hand_error}
 \centering
 \begin{tabular}{|l|l|c|}
  \hline
  \multicolumn{2}{|l|}{Methods }   & Error (mm)   \\
  \hline
  \multicolumn{2}{|l|}{CNNs~\cite{tompson14tog}} & 21.00 \\
  \multicolumn{2}{|l|}{Deep prior~\cite{oberweger15}  }  & 20.00   \\
  \multicolumn{2}{|l|}{Deep model~\cite{Xingyi16ijcai} }   &    16.90   \\
  \multicolumn{2}{|l|}{Feedback loop~\cite{Oberweger15a} }     &  16.50 \\
  \multicolumn{2}{|l|}{Lie-X~\cite{xuchi2016ijcv} } &	14.59	\\
  \multicolumn{2}{|l|}{CNN-baseline}  &	16.13	\\
  \hline
  \multirow{2}{*}{ours}
  &	deep-L2S-para  &  15.84 \\ 
  &	deep-L2S-seq &  \textbf{14.15}  \\ 
  \hline
 \end{tabular}
\end{table}

\begin{figure}
  \centering
  \includegraphics[width=0.475\textwidth]{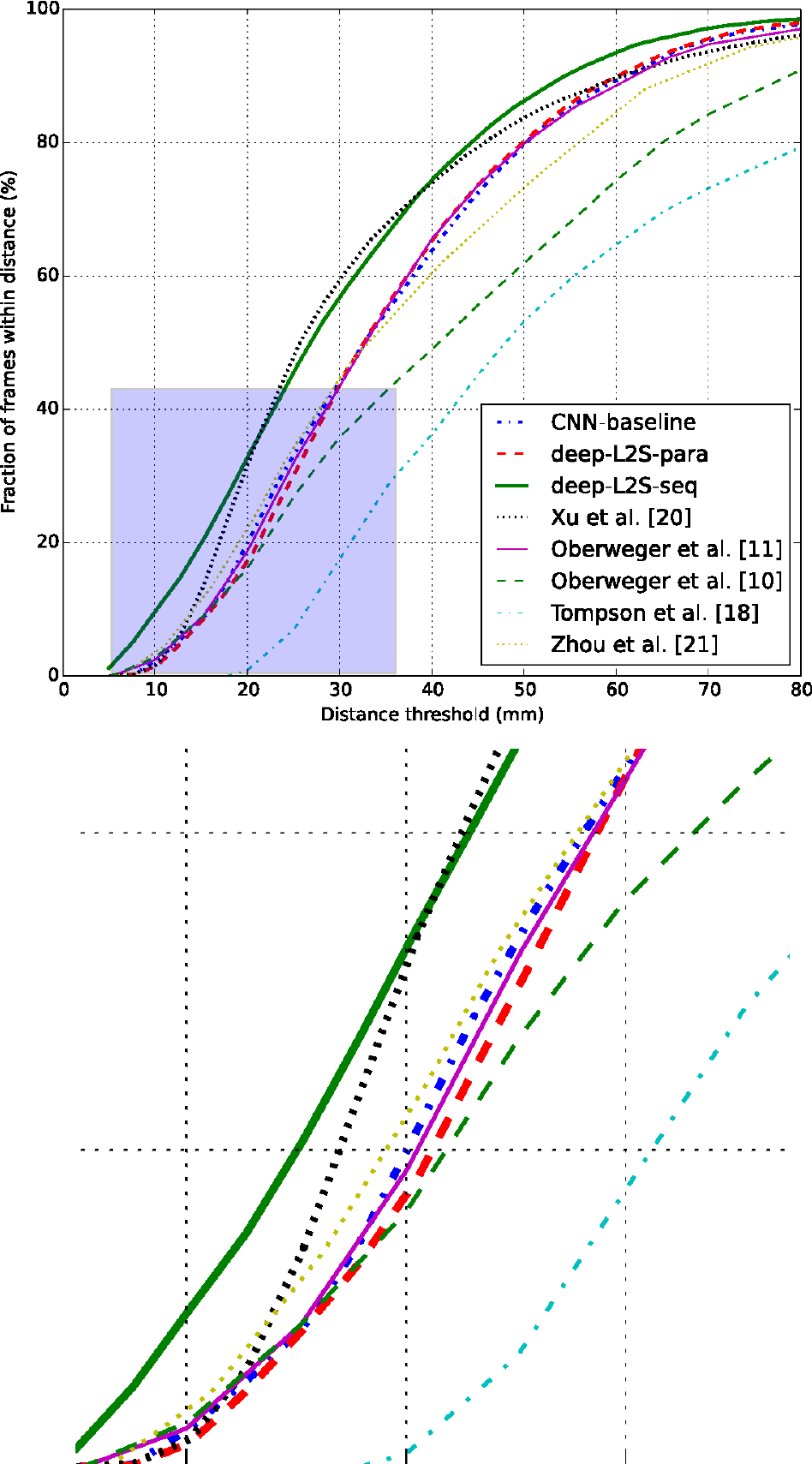}
  \caption{Cumulative error distributions of comparison methods on hand benchmark (i.e. NYU hand dataset~\cite{tompson14tog}). The bottom figure is zoomed in from the blue region in the upper figure. Best viewed in color.}
  \label{fig_hand_cum}
\end{figure}

\begin{figure}
  \centering
  \includegraphics[width=0.48\textwidth]{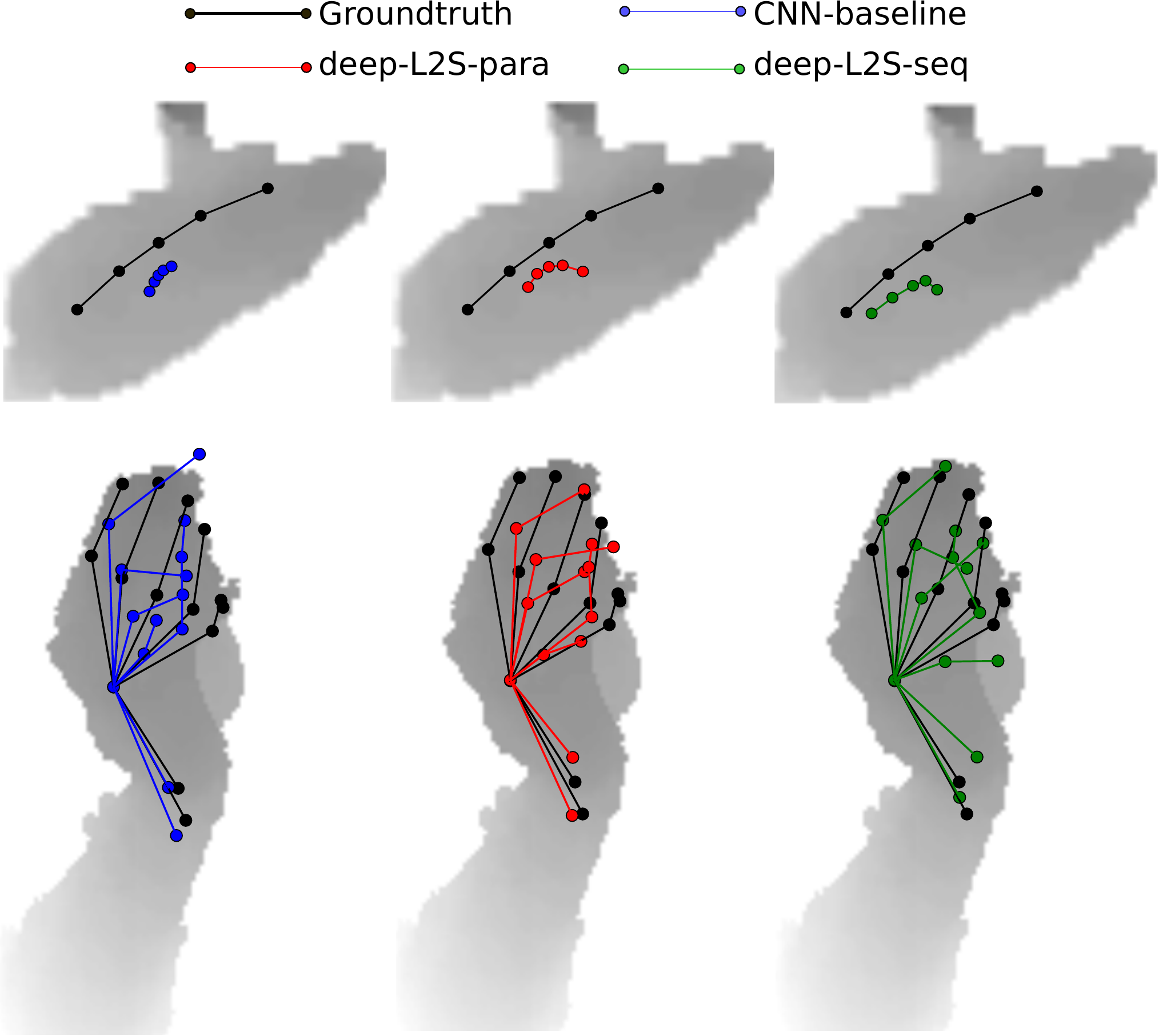}
  \caption{Exemplar failure results on mouse and hand benchmarks. Best viewed in color.}
  \label{fig_fail}
\end{figure}

\begin{figure*}
  \centering
  \includegraphics[width=0.96\textwidth]{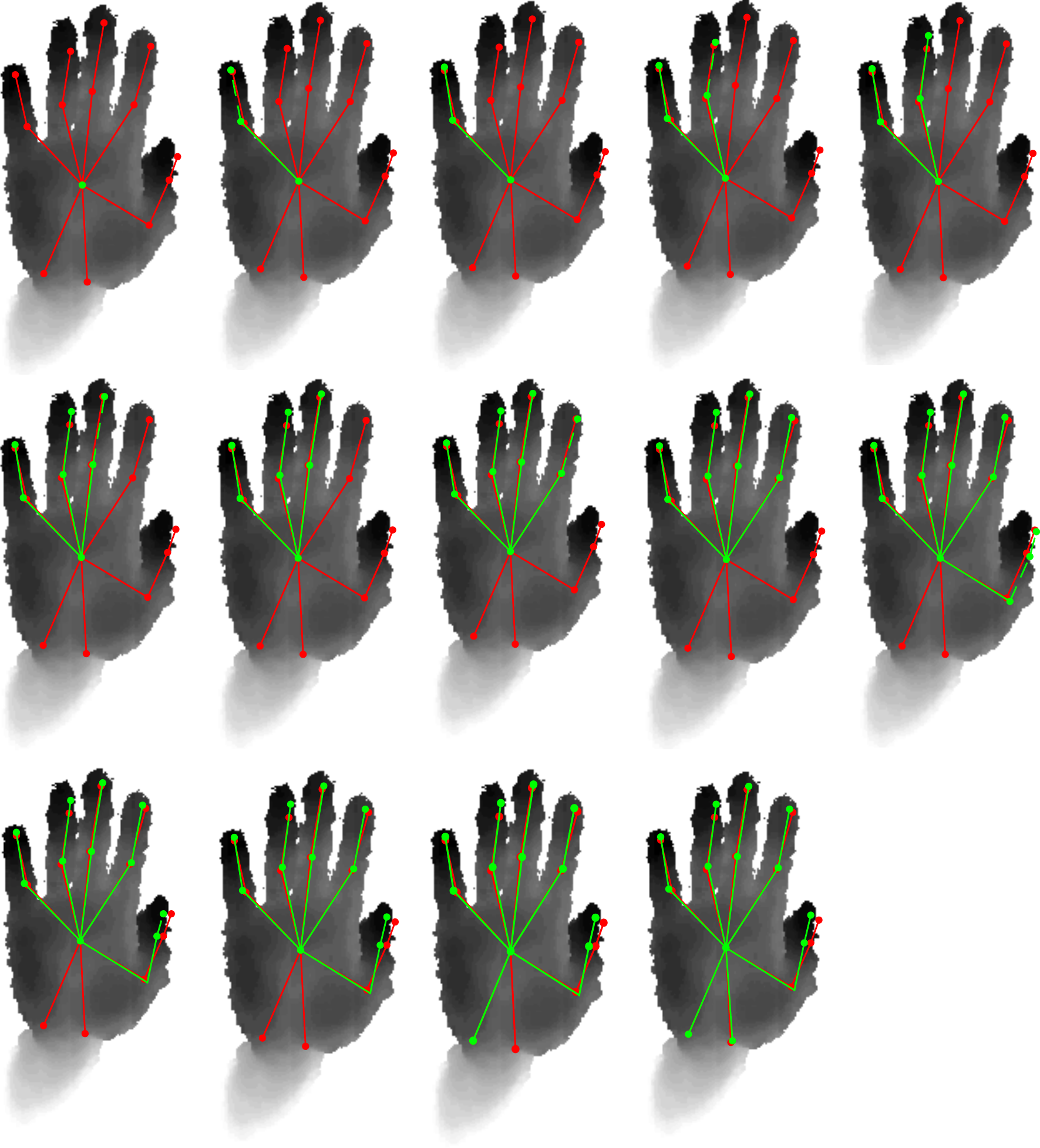}
  \caption{Sequential visualization of the \emph{deep-L2S-seq} method on NYU hand for each joint (from left to right and up to bottom). The red line denotes the results of the \emph{deep-L2S-para} method, which is used as the initialization of the \emph{deep-L2S-seq} method. The green line denotes the result of the \emph{deep-L2S-seq} method, where the solid line denotes the current joint being updated and the dashed line denotes the rest joints awaiting update in the chain.  Best viewed in color.}
  \label{fig_lie_single_visual}
\end{figure*}

\paragraph{Results on human hand benchmark}

Here all methods are compared on the NYU hand dataset that has more complex skeleton model than mouse and fish.
The hand skeleton is represented as a kinematic tree containing multiple chains as depicted in Fig.~\ref{fig_sample_imgs}.
As summarized in Table~\ref{table_hand_error}, our \emph{deep-L2S-seq} continues to outperform the rest comparison methods by a significant margin.
The corresponding cumulative error distributions presented in  Fig.~\ref{fig_hand_cum} reveal more detailed information,
where \emph{deep-L2S-seq} stands out most of the time (i.e over the range of 5-80 mm), except for brief range of around 23-36 mm where \emph{Lie-X} slightly overtakes.
The competitive performance of \emph{deep-L2S-seq} is further verified in the visual comparisons of Fig.~\ref{fig_visual_hand},
where the results of three methods are visualized along with the ground-truth (in black).
%
Compared with the ground-truths, we can clear see the advantages of \emph{deep-L2S-seq} over the other two,
where in different hand images its results are well-conformed to the ground-truths, especially when comparing to the other two methods.

We visualized the process of the \emph{deep-L2S-seq} method in Fig.~\ref{fig_lie_single_visual}. The result of the \emph{deep-L2S-para} method is used as the initial pose of the \emph{deep-L2S-seq} method. Each joint is sequentially updated in the hand skeleton. We can see the clear improvement by the \emph{deep-L2S-seq} method, especially the chain in the  thumb is rectified.

\paragraph{Failure cases}
For the fish benchmark, our approach and especially the \emph{deep-L2S-seq} variant is able to produce very good results.
There are however still big errors occasionally found when working with the mouse and hand datasets, with some displayed as examples in Fig.~\ref{fig_fail}.
For mouse dataset, the failures can often be attribute to the noisy background depth signals that sometimes are wrongly assigned as part of the foregrounds,
which are specifically pronounced for a relatively small object.
For hand dataset, they are usually caused by the occlusion of fingers.

\section{Conclusions}

The problem of 3D pose estimation from single depth images is formulated as structured prediction,
which is further tackled by our approach that involves deep learning with manifold-valued outputs.
Empirically our approach is evaluated on distinct articulated objects of mouse, fish, and human hand, where it is shown to perform competitively with respect to the state-of-the-arts.
Future work includes further investigation of the learning to search paradigm, as well as adaptation of existing algorithms along this line to address pose estimation and related problems.

{\small
\bibliographystyle{ieee}
\bibliography{main}
}

\end{document}